\documentclass{article}

    \PassOptionsToPackage{numbers, compress}{natbib}

\usepackage{siunitx}
\usepackage[utf8]{inputenc} 
\usepackage[T1]{fontenc}    
\usepackage{hyperref}       
\usepackage{url}            
\usepackage{booktabs}       
\usepackage{amsfonts}       
\usepackage{nicefrac}       
\usepackage{microtype}      
\usepackage{xcolor}         
\usepackage{multirow}

\usepackage{graphicx} 
\usepackage{amsmath} 
\usepackage{fancyhdr} 
\usepackage{geometry} 
\usepackage{caption,subcaption}
\usepackage{tikz,graphics,float,epsf}
\usepackage{xcolor}
\usepackage{cancel}
\usepackage{wrapfig}

\usepackage{hyperref}
\hypersetup{colorlinks, citecolor=blue}

\usepackage[english]{babel}
\usepackage{csquotes}
\usepackage{enumitem}

\usepackage{amsfonts} 
\usepackage{natbib}
\usepackage{algorithm}
\usepackage{algpseudocode}
\usepackage{amssymb}
\usepackage{amsthm}
\usepackage{bbold}
\usepackage{amsmath}
\usepackage{mathtools}
\usepackage{xspace}

\theoremstyle{definition}

\theoremstyle{remark}

\theoremstyle{assumption}

\definecolor{es-blue}{rgb}{0,0.4,0.8}

\title{Maneuver Decision-Making Through Proximal Policy Optimization And Monte Carlo Tree Search}

\makeatletter
\newcommand{\printfnsymbol}[1]{%
  \textsuperscript{\@fnsymbol{#1}}%
}

\author{Zhang Hong-Peng\thanks{\fontsize{8}{8} 
		XXXX@XXXX.com.}
}

\begin{document}

\maketitle

\begin{abstract}
Maneuver decision-making can be regarded as a Markov decision process and can be address by reinforcement learning. However, original reinforcement learning algorithms can hardly solve the maneuvering decision-making problem. One reason is that agents use random actions in the early stages of training, which makes it difficult to get rewards and learn how to make effective decisions. To address this issue, a method based on proximal policy optimization and Monte Carlo tree search is proposed. The method uses proximal policy optimization to train the agent, and regards the results of air combat as targets to train the value network. Then, based on the value network and the visit count of each node, Monte Carlo tree search is used to find the actions with more expected returns than random actions, which can improve the training performance. The ablation studies and simulation experiments indicate that agents trained by the proposed method can make different decisions according to different states, which demonstrates that the method can solve the maneuvering decision problem that the original reinforcement learning algorithm cannot solve.

\end{abstract}

\section{Introduction}
\label{sec:intro}

With the development of unmanned aerial vehicle and related technologies, autonomous maneuver decision-making will inevitably become an important direction for unmanned aerial vehicles in the next stage. For example, the artificial intelligence program, Alpha, developed by the University of Cincinnati in the United States defeated human pilots in an air combat simulation system called AFSIM developed by Boeing~\citep{ernest2016genetic}. The Alpha artificial intelligence system adopts the genetic fuzzy tree algorithm~\citep{ernest2016genetic}, which trains and optimizes a tree-like fuzzy inference system to achieve intelligent maneuver decision-making. Therefore, investigating maneuver decision-making methods based on artificial intelligence is of great significance.

Liu et al.~\citep{liu2017deep} studied autonomous maneuver decision-making on the basis of DQN~\citep{mnih2015human}. They divided the maneuvers into five types of maneuvers: top left, bottom left, top right, bottom right, and hold. They used distance and angles as reward functions to achieve attacks on opponent aircraft. He et al.~\citep{he2017air} used Monte Carlo tree search (MCTS) to explore potential actions taken by air combat agents, and selected the action with the greatest advantage of air combat from them. Eloy et al.~\citep{garcia2021differential} used differential game theory to create beyond-visual-range tactics to attack static high-value targets. Wang et al.~\citep{yuan2022research} focused on the problem of insufficient exploration ability of Ornstein-Uhlenbeck strategy and proposed a heuristic reinforcement learning (RL) algorithm with heuristic exploration strategy. Huang et al.~\citep{huang2018maneuvering} proposed a maneuver decision-making method based on approximate dynamic programming and swarm intelligence, and solved the optimal maneuver control quantities with an improved ant lion optimization algorithm for maneuver decision-making. Wang et al.~\citep{wang2021autonomous} proposed an autonomous maneuver strategy of aircraft swarms for beyond-visual-range air combat using deep deterministic policy gradient and validated the effectiveness of the method in multi-scene simulations.

Although RL performs well and even surpasses humans in many tasks~\citep{vinyals2019grandmaster,schrittwieser2020mastering}, it costs much time for training and the training process is unstable. Most importantly, RL may not be able to address some tasks. One reason for the problem is that agents choose random actions in the early stages of training, which makes it difficult to obtain rewards. Therefore, when using RL to address maneuver decision-making, it is difficult for agents to learn to make effective decisions. To address this problem, we proposes a method based on proximal policy optimization (PPO) and MCTS, named PPO-MCTS. The method adopts PPO to train the agent and trains the value network based on the results of air combat in a supervised manner. Then, according to the value network and the visit counts of nodes, MCTS is able to select actions with higher expected returns than random actions, which can balance exploration and exploitation and improve the training efficiency.

\section{Related work}
\label{sec:pre}
We model the maneuver decision-making problem as a Markov decision process~\citep{hester2018deep}. The goal of RL is to maximize expected cumulative rewards~\citep{bertsekas2019reinforcement}, which is an effective method for Markov decision process. Horgan et al.~\citep{horgan2018distributed} proposed a distributed structure for deep RL. This method focuses on using the most important data by prioritized experience replay, and enables agents to effectively learn from more data. To solve decision-making problems in discrete and continuous action spaces, Li et al. proposed the Hyar algorithm~\citep{li2021hyar}. The main idea of Hyar is to establish a unified and decodable representation space for the original mixed discrete continuous actions, convert the learning of the original mixed policy into the continuous policy learning of the hidden action representation space, and then use TD3 for hidden policy learning. Vecerik et al. proposed the DDPGfD algorithm to control real robots~\citep{vecerik2017leveraging}. DDPGfD uses a demonstration of muscle motion to guide the RL algorithm, enabling robots to quickly and safely learn policies to solve operational tasks without the need for handcrafted reward functions. Perolat et al. used the RL algorithm to train agents to become proficient in Western Army chess, known as DeepNash. The core of DeepNash is an MFRL method called regularization Nash dynamics (R-NaD). DeepNash combined R-NaD with deep neural networks to learn a highly competitive policy to find a Nash equilibrium. Jin et al. proposed imitation-relaxation RL to enable quadruped robots to move rapidly and stably~\citep{jin2022high}, which enables a MIT-MiniCheetah-like robot to run stably at a speed of 5 m/s.

MCTS is a best-first search method based on random exploration~\citep{chaslot2008progressive}. Gelly et al. proposed fast action value estimation and heuristic functions to improve the effectiveness of MCTS in computer Go~\citep{gelly2011monte}. Bryan et al. used AlphaFold~\citep{jumper2021highly} to overcome the limitations of predicting protein complexes with 10 to 30 chains and created a graph traversal algorithm based on MCTS to eliminate overlapping interactions, allowing for the gradual assembly of large protein complexes~\citep{bryant2022predicting}. Wang et al. proposed a neural structure search agent based on MCTS, named AlphaX~\citep{wang2019alphax}. It improves search efficiency by adaptively balancing exploration and exploitation, and predicts the accuracy of the network through a meta deep neural network to guide the search towards promising areas. Bai et al. described the relocation of the worst wind turbine as an RL problem and used MCTS combined with evolutionary algorithms to improve the utilization ability of the algorithm~\citep{bai2022wind}, where relocation of worst wind turbines is cast as a reinforcement learning (RL) problem. A searching approach used in Alpha-Go~\citep{silver2016mastering} is adopted to improve the exploitation capability. Mo et al. proposed a RL-based method combined with RL agent and MCTS to reduce unsafe behaviors~\citep{mo2021safe}. This algorithm added an additional reward to risky actions and used MCTS based security policy search. Luo et al. regarded the truss layout problem as a Markov decision process and proposed an algorithm called AlphaTruss based on MCTS to generate the optimal truss layout~\citep{luo2022alphatruss}.

\section{Method}
\label{sec:method}

\subsection{Models} 
The kinematic and dynamic model of the aircraft is shown as follows~\citep{williams2007three}:
\begin{align}
\label{eq:1}
\begin{cases}
\dot{x}=v\cos\gamma\cos\phi\\
\dot{y}=v\cos\gamma\sin\phi\\
\dot{z}=v\sin\gamma\\
\dot{v}=g(n_x-\sin\gamma)\\
\dot{\gamma}=\frac{g}{v}(n_z\cos\mu-\cos\gamma)\\
\dot{\psi}=\frac{g}{v\cos\gamma}n_z\sin\mu\\
\end{cases}
\end{align}

where x, y, and z are the coordinate of the aircraft in the air combat environment. $\gamma$ is the pitch angle, $\phi$ is the yaw angle. v is the velocity, g is the acceleration of gravity. Roll angle , tangential overload , and normal overload  are control parameters. The kinematic model of the missile is :
\begin{align}
\label{eq:2}
\begin{cases}
\dot{x}_m=v_m\cos\gamma_m\cos\phi_m\\
\dot{y}_m=v_m\cos\gamma_m\sin\phi_m\\
\dot{z}_m=v_m\sin\gamma_m\\
\dot{v}_m=\frac{(P_m-Q_m)g}{G_m}-g\sin\gamma_m\\
\dot{\phi}_m=\frac{n_{mc}g}{v_m\cos\gamma_m}\\
\dot{\gamma}_m=\frac{(n_{mh}-\cos\gamma_m)g}{v_m}\\
\end{cases}
\end{align}
where $x_m$, $y_m$, and $z_m$ are three-dimensional coordinates of the missile. $v_m$ represents missile speed, $\gamma_m$ and $\phi_m$ are pitch angle and yaw angle, respectively. $n_{mc}$ and $n_{mh}$ are control signals. $P_m$, $Q_m$ and $G_m$ are thrust, resistance and mass, respectively:
\begin{align}
\label{eq:3-5}
P_m&=
\begin{cases}
P_0 \quad t \leq t_w\\
0 \quad t > t_w
\end{cases}\\
Q_m&=\frac{1}{2}\rho v_m^{2}S_mC_{Dm}\\
G_m&=
\begin{cases}
G_0-G_tt \quad t \leq t_w\\
G_0-G_tt_w \quad t > t_w
\end{cases}
\end{align}
where $t_w = 12.0 s$, $\rho = 0.607$, $Sm = 0.0324$, $C_{Dm} = 0.9$. $P_0$ is the average thrust, $G_0$ is the initial mass, $G_t$ is the rate of flow of fuel. K is the guidance coefficient of proportional guidance law.
\begin{align}
\label{eq:6}
\begin{cases}
n_{mc}&=K\frac{v_m\cos\gamma_t}{g}[\dot{\beta}+\tan\epsilon\tan(\epsilon+\beta)\dot{\epsilon}]\\
n_{mh}&=\frac{v_mK\dot{\epsilon}}{g\cos(\epsilon+\beta)}\\
\beta&=\arctan(r_y/r_x)\\
\epsilon&=\arctan(r_z/\sqrt{(r_x^{2}+r_y^{2})})\\
\dot{\beta}&=(\dot{r}_yr_x-\dot{r}_xr_y)/(r_x^{2}+r_y^{2})\\
\dot{\epsilon}&=\frac{(r_x^{2}+r_y^{2})\dot{r}_z-r_z(\dot{r}_xr_x+\dot{r}_yr_y)}{R^{2}\sqrt{(r_x^{2}+r_y^{2})}}
\end{cases}
\end{align}

\subsection{PPO} 

PPO is a policy gradient based algorithm. The core of the policy gradient algorithms is the policy gradient theorem~\citep{schulman2017proximal}. The goal is to maximize the cumulative reward J by sequentially selecting actions. Then, in order to ensure monotonous improvement of the policy, the author proposes a objectives for optimization. On the basis of~\citep{silver2014deterministic}, a clipped surrogate objective is used in PPO to limit large policy updates, without the need for second-order optimization.

\subsubsection{MCTS}

\begin{figure}[]
	\vspace{-9ex}
	\centering
	\includegraphics[width=\textwidth]{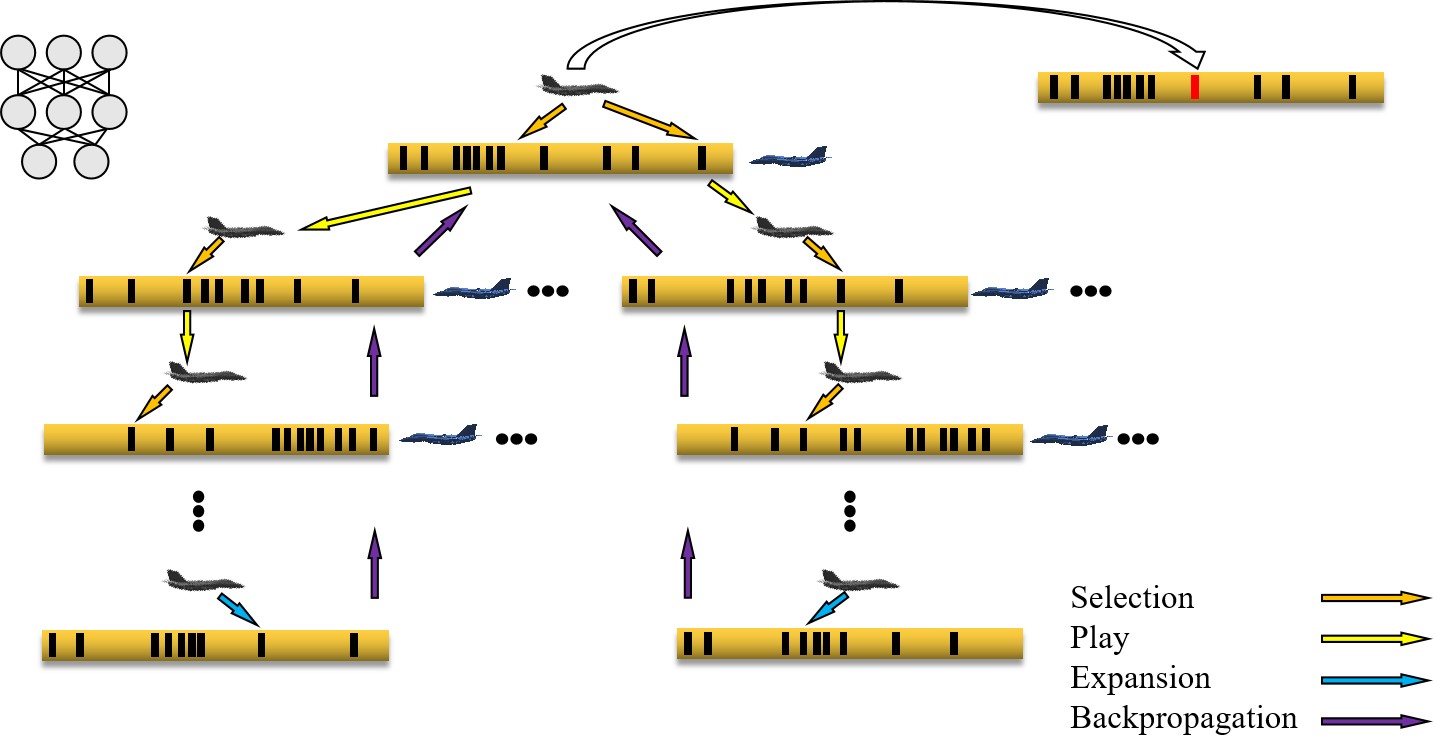}
	\caption{}
	\label{fig:aaa}
\end{figure}

The agent trained by RL outputs only one action per time step when it interacts with the environment. This action comes from a normal distribution, so its randomness usually leads to less returns, which requires more data for training the agent in order to get better performance and often leads to failure. To address this problem, we propose incorporating MCTS into the process of selecting actions in RL, namely, use MCTS to search for actions in the continuous action space to obtain more returns than random actions. Concretely, sample some actions randomly from the normal distribution first. Then, use MCTS to select an action from these actions instead of random selection. Finally, use the action selected by MCTS to interact with the environment. The steps for selecting action by means of MCTS are shown in Figure~\ref{fig:aaa}.

Each node of MCTS stores a set of statistics: {N (s, a), W (s, a), Q (s, a), P (s, a)}. Where, N represents the visit counts of the node, W represents the total value of the node, Q represents the mean value of the node, and P represents the prior probability of the node. The search process of MCTS can be divided into four steps: selection, simulation, extension, and backpropagation. In the selection phase, MCTS selects an action. The method can balance exploration and exploitation, that is, the influence of prior probability and mean value on the selection action is taken into account in the search. In the simulation phase, the selected action in the selection phase is executed until the leaf node is reached. In the expansion phase, the prior probability of each action is calculated and the value network is used to evaluate the value of the action. In the backpropagation phase, the number of visits, total value, and mean value of nodes are updated. The actions output by MCTS have higher value than randomly selected actions, so their results may be better than that of random actions, that is, using these actions may result in more rewards for the agent.

The agent chooses different actions through MCTS given different states, and executes these actions in the air combat environment to arrive at a new state and get corresponding rewards, and these samples composed of states, actions, and rewards are saved. Then, PPO is used to train the agent by these samples to enable the agent to learn policies that can acquire more rewards.

\subsubsection{Air Combat State}

The input of the neural networks is composed of the following vector with min-max normalization: $\phi$, $\gamma$, v, z, d, f1, $\phi$, $\gamma$, $\phi$1, $\gamma$1, d1, $\beta$, f2. The neural networks have two hidden layers and the activation function is hyperbolic tangent. Finally, the output layer includes four quantities. The first three are normal overload, tangential overload and roll angle. The last one is whether to launch the missile.

\section{Experiments} 
\label{sec:exp}

In order to investigate the performance of the proposed method, this section verifies the training results of PPO-MCTS and the maneuver ability of trained agents by means of ablation studies and simulation experiments, respectively. The number of actions is 9 for each MCTS with 20 simulations. Three independent experiments for PPO-MCTS and PPO are conducted, and the experimental results are recorded. After each iteration of training, the current agent is saved and 36 previous agents are randomly selected. Then, three air combat simulations between the current agent and the previous agent are performed, and the simulation results are recorded. The experimental hyperparameters are listed in Table~\ref{tab:hyper}.

\begin{table}[]
	\centering
	\caption{Hyperparameters}
	\label{tab:hyper}
	\begin{tabular}{|c|c|}
		\hline
		\textbf{Name  }     & \textbf{Value}  \\ \hline
		Velocity  & $\left[250~\si{\meter\per\second},400~\si{\meter\per\second}\right]$   \\ \hline
		Batchsize                 & 1024   \\ \hline
		Optimizer                 & Adam  \\ \hline
		Actor learning rate                 & 0.002  \\ \hline
		Critic learning rate                 & 0.001  \\ \hline
		Actor architecture            & 256*256*4  \\ \hline
		Critic architecture         & 256*256*1  \\ \hline
		Activate function                 & tanh  \\ \hline
		Epoach                 & 6  \\ \hline
		$\gamma$                           & 0.99  \\ \hline
	\end{tabular}
\end{table}

\subsection{Ablation Studies}

\begin{figure}[]
	\vspace{-9ex}
	\centering
	\begin{subfigure}[b]{0.5\textwidth}
		\centering
		\includegraphics[width=\textwidth]{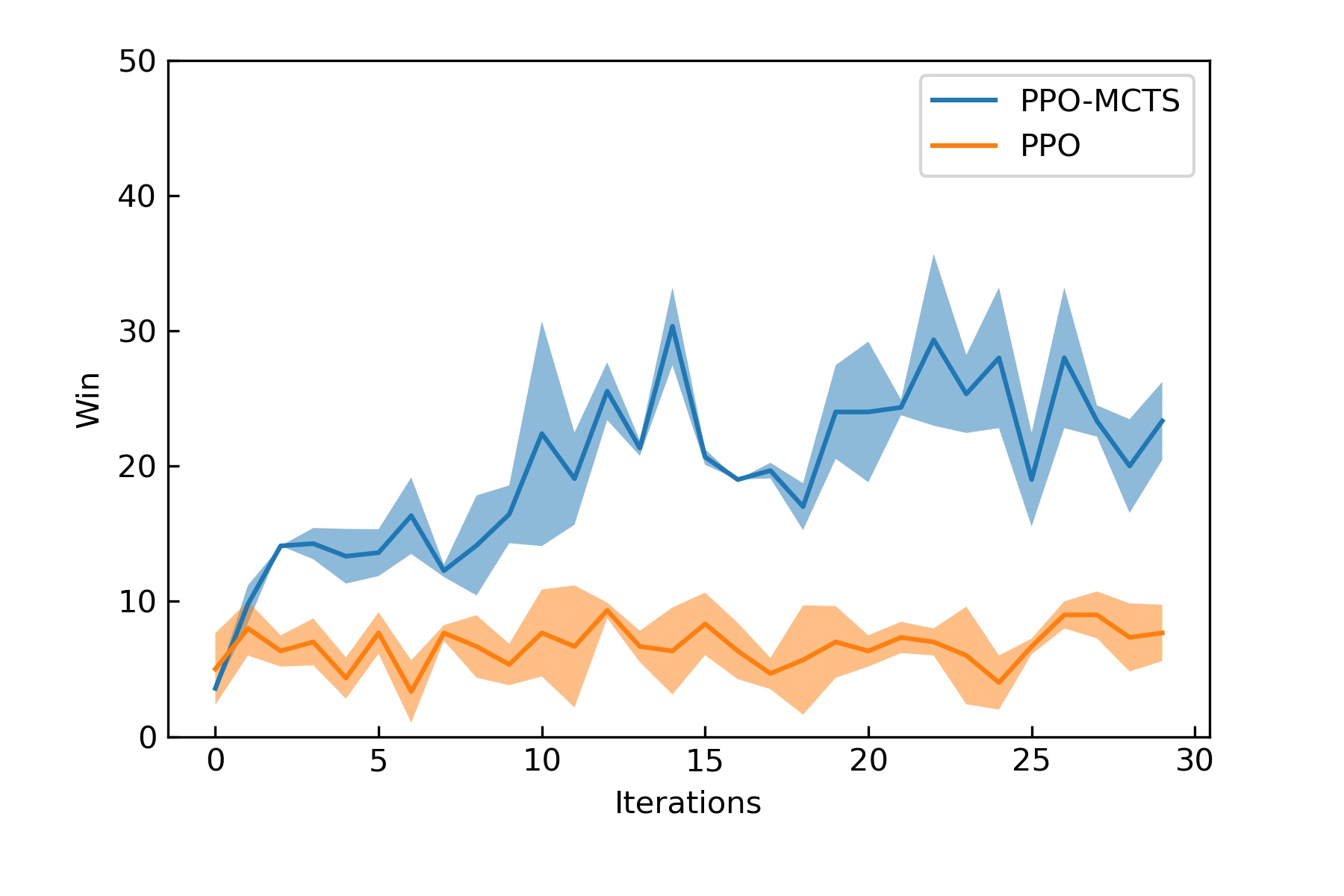}
		\caption{}
		\label{fig:win}
	\end{subfigure}
	\vfill
	\begin{subfigure}[b]{0.5\textwidth}
		\centering
		\includegraphics[width=\textwidth]{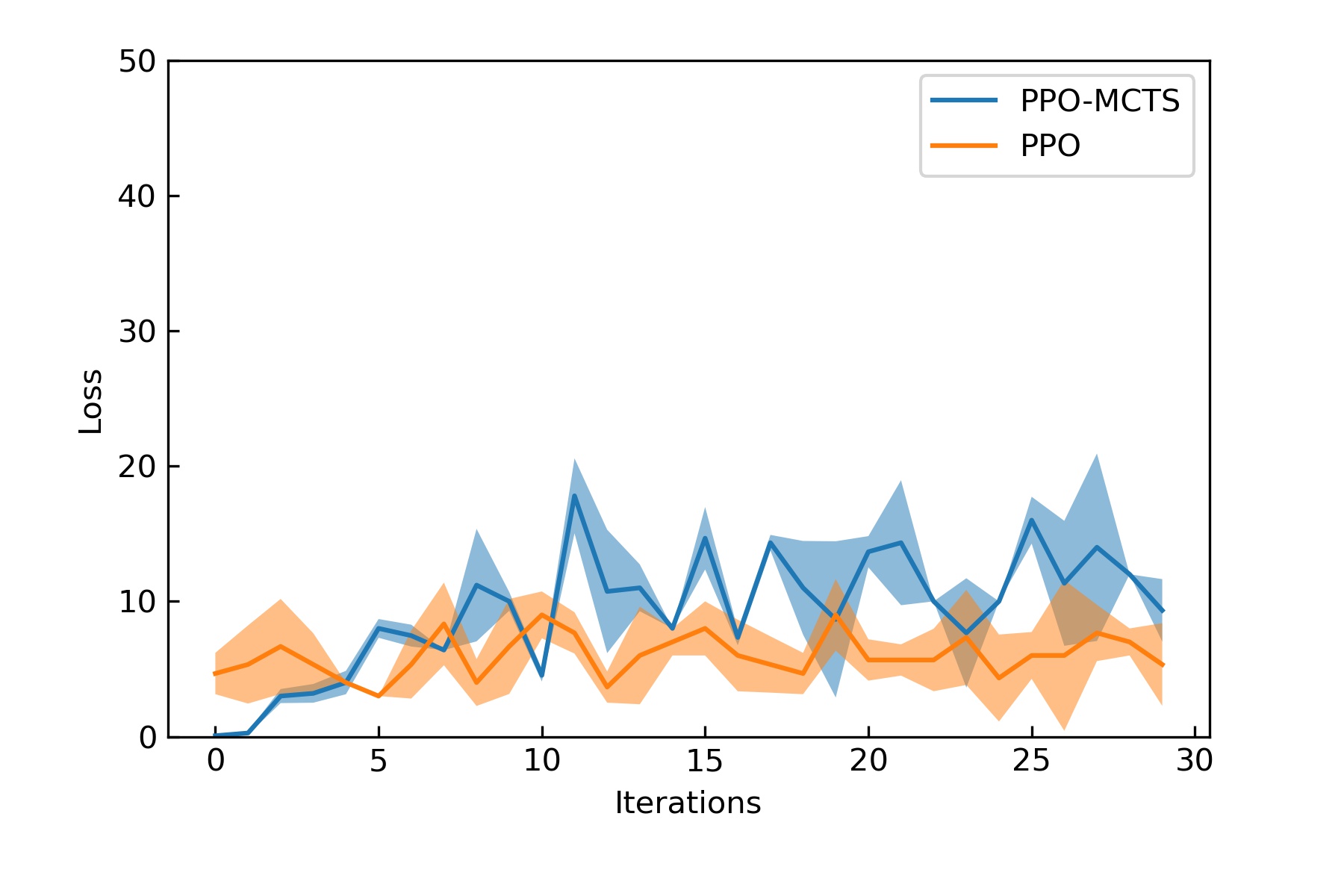}
		\caption{}
		\label{fig:loss}
	\end{subfigure}
	\vfill
	\begin{subfigure}[b]{0.5\textwidth}
		\centering
		\includegraphics[width=\textwidth]{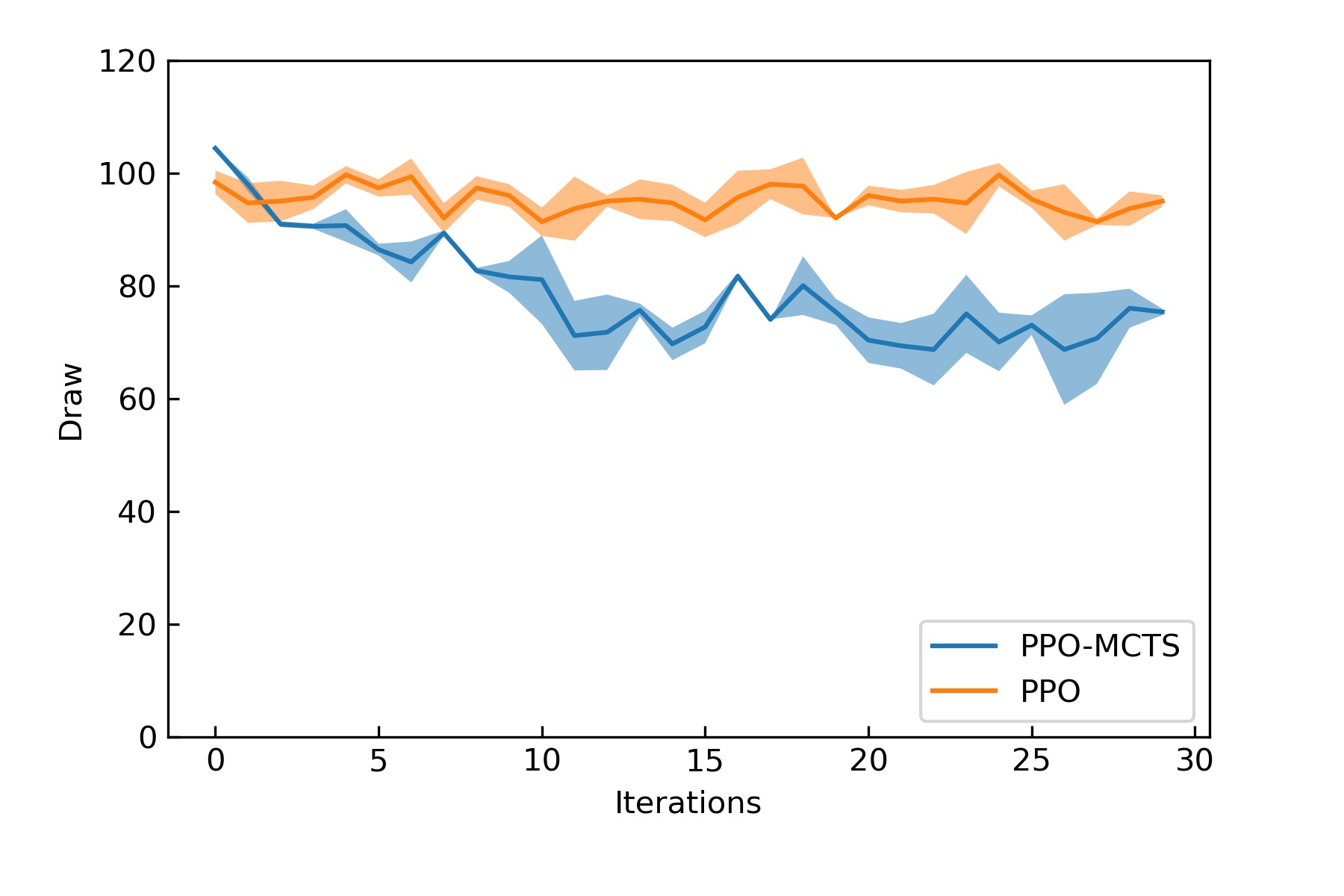}
		\caption{}
		\label{fig:draw}
	\end{subfigure}
	\vfill
	\caption{\textbf{(a)} Win. \textbf{(b)} Loss. \textbf{(c)} Draw.}
	\vspace{-4ex}
	\label{fig:train}
\end{figure}

Figure~\ref{fig:train} demonstrates the number of win, loss, and draw of agents trained with PPO-MCTS and PPO in each test. Among them, Figure~\ref{fig:win} represents the number of win, Figure 2b represents the number of loss, Figure 2c represents the number of draw. The solid line represents the mean, and the shaded part represents the corresponding standard deviation. From Figure 2a, it can be seen that the number of win of PPO-MCTS gradually increases, but the number of win of PPO does not increase at all, which indicates that PPO-MCTS is valid but PPO is invalid. Similarly, as shown in Figure 2b, more losses indicates that past agents can make effective decisions, that is, the method with more losses is more effective. As shown Figure 2c, the number of draw of PPO-MCTS gradually decreases, indicating that agents trained by PPO-MCTS gradually acquires effective ability of decision.

\section{Discussion}
\label{sec:Discussion}
The purpose of maneuver decision-making is to defeat the opponent, or to evade the attack of the opponent's attack. Maneuver decision-making is regarded as a Markov decision process in the article and RL is used to solve maneuver decision-making problems. Due to the difficulty of maneuvering decision-making, the original PPO algorithm cannot address this problem. Therefore, we propose to combine PPO with MCTS to solve maneuvering decision-making problems. MCTS is an effective forward planning algorithm that can be used to select actions. In the ablation studies, both PPO and PPO-MCTS are used to train agents. The number of wins corresponding to PPO is always only a few, while the number of wins corresponding to PPO-MCTS gradually increases. The results indicate that agents trained by PPO-MCTS can make effective decisions, because more wins represent more times that targets hit by missiles and ineffective maneuvers can hardly make the missile hit the target. In the simulation experiments, we present the simulation results of untrained agents and trained agents. According to the results, random decisions are invalid obviously. Because even at an advantage, the untrained agent is not able to win. On the contrary, the trained agent can make effective maneuvers. For example, even if the target is behind it, it changes the flight direction and launches the missile, which cannot be accomplished through random maneuvers.

\section{Conclusion}
\label{sec:Conclusion}
RL is adopted for maneuver decision-making problem in this article. Due to the ineffectiveness of the original PPO algorithm for maneuver decision-making, we propose PPO-MCTS. The method integrates MCTS into the process of selecting actions in PPO, in which MCTS is used to search for actions in the continuous action space with more returns than that of random actions. Then, PPO is used to train agents in order to enable them to make effective maneuver decision-makings given different states. The effectiveness of the proposed method is verified through ablation studies and simulation experiments. The results of ablation studies indicate that PPO-MCTS can increase the number of win, while PPO without MCTS cannot increase the number of win. The results of simulation experiments indicate that PPO-MCTS can gradually enable agents to learn how to make effective decisions. Concretely, at the beginning of the training, the agent makes random and meaningless decisions, and the number of win is less. After the training process, the agent can attack the target located at the rear, and the number of win is more. 

\subsubsection*{Acknowledgments}
\vspace{-0.5em}
\small{
We thank Herzog.}

\normalsize
\bibliography{main.bib}

\end{document}